\documentclass[11pt,a4paper]{article}
\usepackage{authblk}
\usepackage[hyperref]{emnlp-ijcnlp-2019}
\usepackage{times}
\usepackage{latexsym}
\usepackage{amsmath}
\usepackage{enumitem}
\usepackage{multirow}
\usepackage{multicol}
\usepackage{tabularx}
\usepackage{subcaption}
\usepackage{amssymb}
\usepackage{booktabs}
\usepackage{array}
\usepackage[pdftex]{graphicx}
\usepackage{url}
\usepackage{xcolor}

\graphicspath{{./figures/}}

\aclfinalcopy % Uncomment this line for the final submission

\allowdisplaybreaks
\DeclareMathOperator*{\softmax}{softmax}
\DeclareMathOperator*{\avg}{avg}

\title{Connecting the Dots: Document-level Neural Relation Extraction \\ with Edge-oriented Graphs}

\author[1]{Fenia Christopoulou}
\author[2,3]{Makoto Miwa}
\author[1]{Sophia Ananiadou}
\affil[1]{National Centre for Text Mining, \authorcr
\textnormal{\normalsize School of Computer Science, The University of Manchester, United Kingdom}}
\affil[2]{Toyota Technological Institute, Nagoya, 468-8511, Japan}
\affil[3]{Artificial Intelligence Research Center (AIRC), \authorcr
\textnormal{\normalsize National Institute of Advanced Industrial Science and Technology (AIST), Japan}}
\affil[ ]{\tt \{efstathia.christopoulou, sophia.ananiadou\}@manchester.ac.uk}
\affil[ ]{\tt makoto-miwa@toyota-ti.ac.jp}

\date{}
\begin{document}
\maketitle
\begin{abstract}
    Document-level relation extraction is a complex human process that requires logical inference to extract relationships between named entities in text. 
    Existing approaches use graph-based neural models with words as nodes and edges as relations between them, to encode relations across sentences. These models are node-based, i.e., they form pair representations based solely on the two target node representations. However, entity relations can be better expressed through unique edge representations formed as paths between  nodes. We thus propose an edge-oriented graph neural model for document-level relation extraction. The model utilises different types of nodes and edges to create a document-level graph. An inference mechanism on the graph edges enables to learn intra- and inter-sentence relations using multi-instance learning internally. Experiments on two document-level biomedical datasets for chemical-disease and gene-disease associations show the usefulness of the proposed edge-oriented approach.\footnote{Source code available at \url{https://github.com/fenchri/edge-oriented-graph}}
\end{abstract}

%%% INTRO
\section{Introduction}
The extraction of relations between named entities in text, known as Relation Extraction (RE), is an important task of Natural Language Processing (NLP). Lately, RE has attracted a lot of attention from the field, in an effort to improve the inference capability of current methods~\cite{zeng2017paths,christopoulou2018walk,luan2019general}. 

In real-world scenarios, a large amount of relations are expressed across sentences. The task of identifying these relations is named inter-sentence RE. Typically, inter-sentence relations occur in textual snippets with several sentences, such as documents.
In these snippets, each entity is usually repeated with the same phrases or aliases, the occurrences of which are often named entity mentions and regarded as instances of the entity. 
The multiple mentions of the target entities in different sentences can be useful for the identification of inter-sentential relations, as these relations may depend on the interactions of their mentions with other entities in the same document.

\begin{figure}[t]
	\centering
	\includegraphics[width=0.49\textwidth]{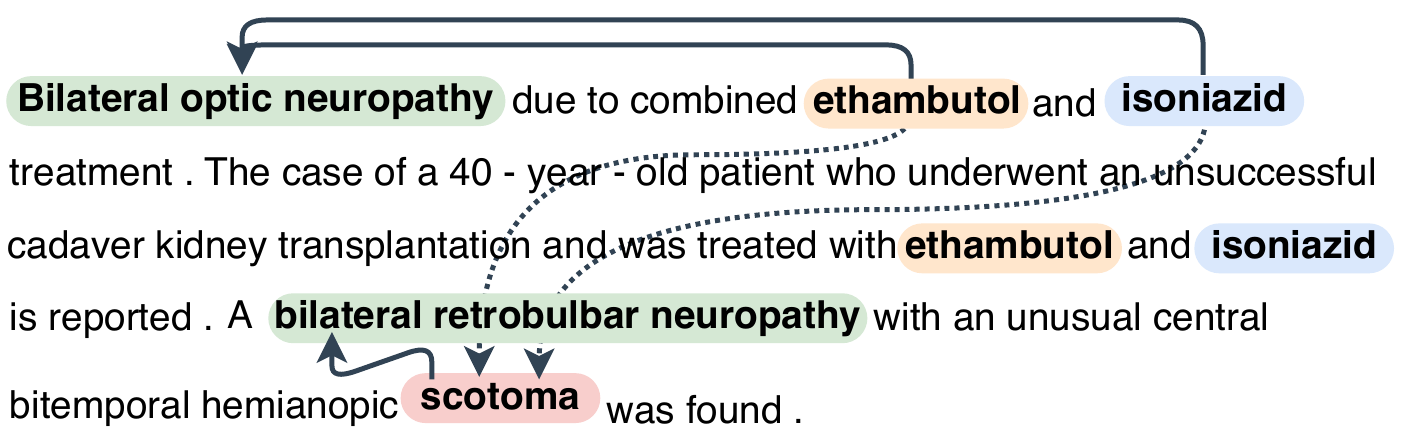}
	\caption{Example of document-level, inter-sentence relations adapted from the CDR dataset \cite{li2016biocreative}. The solid and dotted lines represent intra- and inter-sentence relations, respectively.}
	\label{fig:example}
\end{figure}

As shown in the example of Figure~\ref{fig:example}, the entities \textit{bilateral optic neuropathy}, \textit{ethambutol} and \textit{isoniazid} have two mentions each, while the entity \textit{scotoma} has one mention.
The relation between the chemical \textit{ethambutol} and the disease \textit{scotoma} is clearly inter-sentential. Their association can only be determined if we consider the interactions between the mentions of these entities in different sentences. A mention of \textit{bilateral optic neuropathy} interacts with a mention of \textit{ethambutol} in the first sentence. Another mention of the former interacts with the mention of \textit{scotoma} in the third sentence. This chain of interactions can help us infer that the entity \textit{ethambutol} has a relation with the entity \textit{scotoma}.

The most common technique that is currently used to deal with multiple mentions of named entities is Multi-Instance Learning (MIL). Initially, MIL was introduced by \citet{riedel2010modeling} in order to reduce noise in distantly supervised corpora~\cite{mintz2009distant}. 
In DS, training instances are created from large, raw corpora using Knowledge Base (KB) entity linking and automatic annotation with heuristic rules.
MIL in this setting considers multiple sentences (bags) that contain a pair of entities serving as the multiple instances of this pair.
\citet{verga2018simultaneously} introduced another MIL setting for relation extraction between named entities in a document. 
In this setting, entities mapped to the same KB ID are considered as mentions of an entity concept and pairs of mentions correspond to the pair's multiple instances.
However, document-level RE is not common in the general domain, as the entity types of interest can often be found in the same sentence~\cite{banko2007open}. On the contrary, in the biomedical domain, document-level relations are particularly important given the numerous aliases that biomedical entities can have~\cite{quirk2017distant}.

To deal with document-level RE, recent approaches assume that only two mentions of the target entities reside in the document~\cite{nguyen2018convolutional,verga2018simultaneously} or utilise different models for intra- and inter-sentence RE~\cite{gu2016chemical,li2016cidextractor,gu2017chemical}.  
In contrast with approaches that employ sequential models~\cite{nguyen2018convolutional,gu2017chemical,zhou2016exploiting}, graph-based neural approaches have proven useful in encoding long-distance, inter-sentential information~\cite{peng2017cross,quirk2017distant,gupta2019neural}. These models interpret words as nodes and connections between them as edges. They typically perform on the nodes by updating the representations during training. 
However, a relation between two entities depends on different contexts. It could thus be better expressed with an edge connection that is unique for the pair. 
A straightforward way to address this is to create graph-based models that rely on edge representations rather focusing on node representations, which are shared between multiple entity pairs.

In this work, we tackle document-level, intra- and inter-sentence RE using MIL with a graph-based neural model. Our objective is to infer the relation between two entities by exploiting other interactions in the document. 
We construct a document graph with heterogeneous types of nodes and edges to better capture different dependencies between nodes.
In the proposed graph, a node corresponds to either entities, mentions, or sentences, instead of words. 
We connect distinct nodes based on simple heuristic rules and generate different edge representations for the connected nodes. 
To achieve our objective, we design the model to be edge-oriented in a sense that it learns edge representations (between the graph nodes) rather than node representations.
An iterative algorithm over the graph edges is used to model dependencies between the nodes in the form of edge representations. The intra- and inter-sentence entity relations are predicted by employing these edges.
Our contributions can be summarised as follows:
\begin{itemize}[nolistsep,leftmargin=*]
    \item We propose a novel edge-oriented graph neural model for document-level relation extraction. The model deviates from existing graph models as it focuses on constructing unique nodes and edges, encoding information into edge representations rather than node representations. 
    \item The proposed model is independent of syntactic dependency tools and can achieve state-of-the-art performance on a manually annotated, document-level chemical-disease interaction dataset. 
    \item Analysis of the model components indicates that the document-level graph can effectively encode document-level dependencies. 
    Additionally, we show that inter-sentence associations can be beneficial for the detection of intra-sentence relations.
\end{itemize}

%% Model
\section{Proposed Model}

\begin{figure*}[t!]
    \centering
    \includegraphics[width=\textwidth]{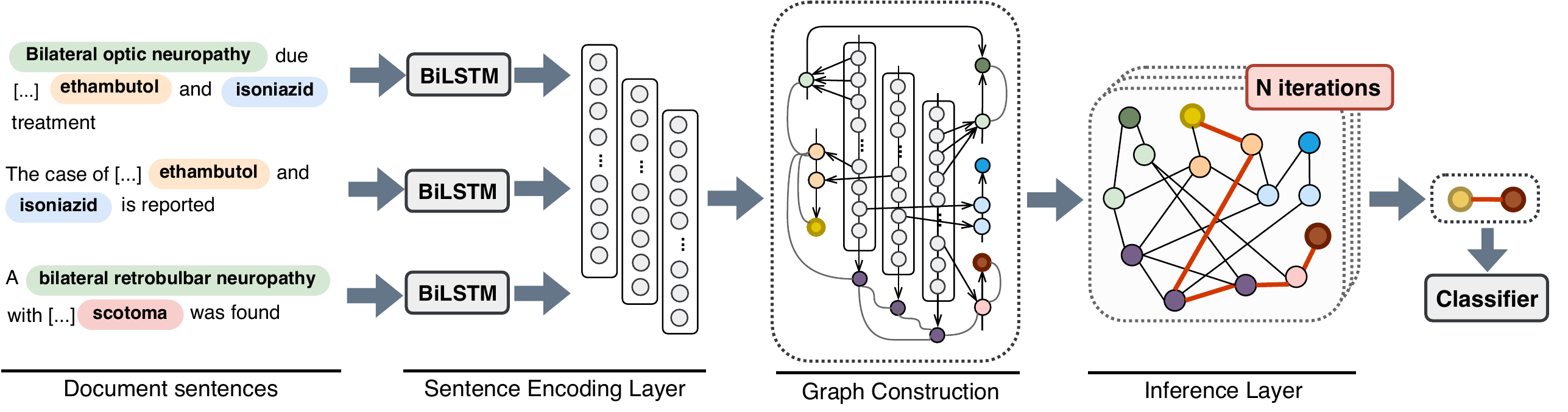}
    \caption{Abstract architecture of the proposed approach. The model receives a document and encodes each sentence separately. A document-level graph is constructed and fed into an iterative algorithm to generate edge representations between the target entity nodes. Some node connections are not shown for brevity.}
    \label{fig:network}
\end{figure*}

We build our model as a significant extension of our previously proposed sentence-level model~\cite{christopoulou2018walk} for document-level RE. The most critical difference between the two models is the introduction and construction of a partially-connected document graph, instead of a fully-connected sentence-level graph. Additionally, the document graph consists of heterogeneous types of nodes and edges in comparison with the sentence-level graph that contains only entity-nodes and single edge types among them. Furthermore, the proposed approach utilises multi-instance learning when mention-level annotations are available.

As illustrated in Figure~\ref{fig:network}, the proposed model consists of four layers: sentence encoding, graph construction, inference and classification layers.
The model receives as input a document with identified concept-level entities and their textual mentions. Next, a document-level graph with multiple types of nodes and edges is constructed. An inference algorithm is applied on the graph edges to generate concept-level pair representations. In the final layer, the edge representations between the target concept-entity nodes are classified into relation categories.

For the remainder of this section, we first briefly introduce the document-level RE task setting and then explain the four layers of the proposed model.

\subsection{Task Setting} 

In concept, document-level RE the input is considered an annotated document. The annotations include concept-level entities (with assigned KB IDs), as well as multiple occurrences of each entity under the same phrase of alias, i.e., entity mentions. We consider the associations of mentions to concept entities given (also known as entity linking~\cite{shen2014entity}). The objective of the task is given an annotated document, to identify all the related concept-level pairs in that document. In this work, we refer to concept-level annotations as \textit{entities} and mention-level annotations as \textit{mentions}.

\subsection{Sentence Encoding Layer}

First, each word in the sentences of the input document is transformed into a dense vector representation, i.e., a word embedding. The vectorised words of each sentence are then fed into a Bi-directional LSTM network (BiLSTM)~\cite{hochreiter1997long,schuster1997bidirectional}, named the encoder.
The output of the encoder results in contextualised representations for each word of the input sentence.

\subsection{Graph Layer}

The contextualised word representations from the encoder are used to construct a document-level graph structure. The graph layer comprises of two sub-layers, a node construction layer and an edge construction layer. We compose the representations of the graph nodes in the first sub-layer and the representations of the edges in the second.

%%% NODES
\subsubsection{Node construction}

We form three distinct types of nodes in the graph: mention nodes (M) $n_m$, entity nodes (E) $n_e$, and sentence nodes (S) $n_s$. 
Each node representation is computed as the average of the embeddings of different elements.
Firstly, mention nodes correspond to different mentions of entities in the input document. The representation of a mention node is formed as the average of the words ($w$) that the mention contains. 
Secondly, entity nodes represent unique entity concepts. The representation of an entity node is computed as the average of the mention ($m$) representations associated with the entity.
Finally, sentence nodes correspond to sentences. A sentence node is represented as the average of the word representations in the sentence.
In order to distinguish different node types in the graph, we concatenate a node type ($t$) embedding to each node representation. 
The final node representations are then estimated as 
$\mathbf{n}_m = [ \avg_{w_i \in m} (\mathbf{w}_i); \mathbf{t}_m]$, 
$\mathbf{n}_e = [ \avg_{m_i \in e} (\mathbf{m}_i); \mathbf{t}_e]$, 
$\mathbf{n}_s = [ \avg_{w_i \in s} (\mathbf{w}_i); \mathbf{t}_s].$

%%% EDGES
\subsubsection{Edge construction}
\label{ssec:edge}

We initially construct non-directed edges between the graph nodes using heuristic rules that stem from the natural associations between the elements of a document, i.e., mentions, entities and sentences. As we cannot know in advance if two entities are related, we do not directly connect entity nodes. Connections between nodes are based on pre-defined document-level interactions. The model objective is to generate entity-to-entity (EE) edge representations using other existing edges in the graph and consequently infer entity-to-entity relations.
The different pre-defined edge types are described below.

\noindent \textbf{Mention-Mention (MM)}: Co-occurrence of mentions in a sentence might be a weak indication of an interaction. For this reason, we create mention-to-mention edges only if the corresponding mentions reside in the same sentence. 
The edge representation between each mention pair $m_i$ and $m_j$ is generated by concatenating the representations of the nodes, the contexts $c_{m_i, m_j}$ and a distance embedding associated with the distance between the two mentions $d_{m_i, m_j}$, in terms of intermediate words: $\mathbf{x}_{\text{MM}} = [ \mathbf{n}_{m_i} ; \mathbf{n}_{m_j} ; \mathbf{c}_{m_i, m_j} ; \mathbf{d}_{m_i, m_j} ]$.
Here, we generate the context representation for these pairs in order to encode local, pair-centric information. We use an argument-based attention mechanism~\cite{wang2016relation}, to measure the importance of other words in the sentence towards the mention, denoting $k \in \{1, 2\}$ as the mention arguments.
\begin{equation}
    \begin{aligned}
        \alpha_{k,i} &= \mathbf{n}_{m_k}^\intercal \mathbf{w}_i,  \\
        \text{a}_{k,i} &= \frac{\exp{( \alpha_{k,i} )}}{\sum_{\substack{j \in [1, n], j \not\in m_k}} \exp{( \alpha_{k, j} )}},  \\        
        \text{a}_i &= (\text{a}_{1,i} + \text{a}_{2,i})/ 2, \\
        \mathbf{c}_{m_1,m_2} &= {\mathbf{H}}^\intercal \; \mathbf{a},
    \end{aligned}
\end{equation}
where $\mathbf{n}_{m_k}$ is a mention node representation, $\mathbf{w}_i$ is a sentence word representation, $\text{a}_i$ is the attention weight of word $i$ for mention pair $m_1, m_2$, $\mathbf{H} \in \mathbb{R}^{w \times d}$ is a sentence word representations matrix, $\mathbf{a} \in \mathbb{R}^{w}$ is the attention weights vector for the pair and $\mathbf{c}_{m_1,m_2}$ is the final context representation for the mention pair.

\noindent \textbf{Mention-Sentence (MS)}: Mention-to-sentence nodes are connected only if the mention resides in the sentence. Their initial edge representation is constructed as a concatenation of the mention and sentence nodes, $\mathbf{x}_{\text{MS}} = [\mathbf{n}_m ; \mathbf{n}_s]$.

\noindent \textbf{Mention-Entity (ME)}: We connect a mention node to an entity node if the mention is associated with the entity, $\mathbf{x}_{\text{ME}} = [\mathbf{n}_m ; \mathbf{n}_e]$.

\noindent \textbf{Sentence-Sentence (SS)}: Motivated by \citet{quirk2017distant}, we connect sentence nodes to encode non-local information. The main differences with prior work is that our edges are unlabelled, non-directed and span multiple sentences. 
To encode the distance between sentences, we concatenate to the sentence node representations their distance in the form of an embedding: $\mathbf{x}_{\text{SS}} = [ \mathbf{n}_{s_i} ; \mathbf{n}_{s_j} ; \mathbf{d}_{s_i, s_j} ]$. 
We connect all sentence nodes in the graph. We consider \textbf{$\text{SS}_{\text{direct}}$} as direct, ordered edges (distance equal to $1$) and \textbf{$\text{SS}_{\text{indirect}}$} as indirect, non-ordered edges (distance $> 1$) between S nodes, respectively. In our setting, \textbf{$\text{SS}$} denotes the combination of  $\text{SS}_{\text{direct}}$ and $\text{SS}_{\text{indirect}}$. 

\noindent \textbf{Entity-Sentence (ES)}: To directly model entity-to-sentence associations, we connect an entity node to a sentence node if at least one mention of the entity resides in this sentence, $\mathbf{x}_{\text{ES}} = [\mathbf{n}_e ; \mathbf{n}_s]$.

In order to result in edge representations of equal dimensionality, we use different linear reduction layers for different edge representations,
\begin{equation}
    \mathbf{e}^{(1)}_{z} = \mathbf{W}_z \; \mathbf{x}_z,
    \label{eq:linear}
\end{equation}
where $\mathbf{e}^{(1)}_{z}$ is an edge representation of length $1$, $\mathbf{W}_z \in \mathbb{R}^{d_z \times d}$ corresponds to a learned matrix and $z \in [ \text{MM},  \text{MS}, \text{ME}, \text{SS}, \text{ES} ]$.

\subsection{Inference Layer}
\label{sec:infenrence}

We utilise an iterative algorithm to generate edges between different nodes in the graph, as well as to update existing edges.
We initialise the graph only with the edges described in Section~\ref{ssec:edge}, meaning that direct entity-to-entity (EE) edges are absent. 
We can only generate EE edge representations by representing a path between their nodes.
This implies that entities can be associated through an edge path of minimum length equal to $3$\footnote{Length $2$ for an intra-sentence pair (E-S-E) or length $3$ for an inter-sentence pair (E-S-S-E)}. 

For this purpose, we adapt our two-step inference mechanism, proposed in \citet{christopoulou2018walk}, to encode interactions between nodes and edges in the graph and hence model EE associations. 

At the first step, we aim to generate a path between two nodes $i$ and $j$ using intermediate nodes $k$. We thus combine the representations of two consecutive edges $e_{ik}$ and $e_{kj}$, using a modified bilinear transformation. 
This action generates an edge representation of double length. 
We combine all existing paths between $i$ and $j$ through $k$. The $i$, $j$, and $k$ nodes can be any of the three node types E, M, or S. Intermediate nodes without adjacent edges to the target nodes are ignored.
\begin{equation}
   f \left( \mathbf{e}_{ik}^{(l)}, \mathbf{e}_{kj}^{(l)} \right) = \sigma \left( \mathbf{e}_{ik}^{(l)} \odot \left( \mathbf{W} \; \mathbf{e}_{kj}^{(l)}   \right) \right), 
   \label{eq:generate}
\end{equation}
where $\sigma$ is the sigmoid non-linear function, $\mathbf{W} \in \mathbb{R}^{d_z \times d_z}$ is a learned parameter matrix, $\odot$ refers to element-wise multiplication, $l$ is the length of the edge and $\mathbf{e}_{ik}$ corresponds to the representation of the edge between nodes $i$ and $k$.

During the second step, we aggregate the original (short) edge representation and the new (longer) edge representation resulted from Equation (\ref{eq:generate}) with linear interpolation as follows:
\begin{equation}
    \textbf{e}_{ij}^{(2l)} = \beta \; \textbf{e}_{ij}^{(l)} + (1-\beta) \; \sum_{k \neq i,j} f \left( \mathbf{e}_{ik}^{(l)}, \mathbf{e}_{kj}^{(l)} \right),
    \label{eq:aggregate}
\end{equation}
where $\beta \in [0, 1]$ is a scalar that controls the contribution of the shorter edge presentation. In general $\beta$ is larger for shorter edges as we expect that the relation between two nodes is better expressed through the shortest path between them~\cite{xu2015classifying,Borgwardt2005shortest}.

The two steps are repeated a finite number of times $N$. The number of iterations is correlated with the final length of the edge representations. With initial edge length $l$ equal to $1$, the first iteration results in edges of length up-to $2$. The second iteration results in edges of length up-to $4$. Similarly, after $N$ iterations, the length of edges will be up-to $2^{N}$.

\subsection{Classification Layer}
\label{sec:classification}

To classify the concept-level entity pairs of interest, we incorporate a softmax classifier, using the entity-to-entity edges (EE) of the document graph that correspond to the concept-level entity pairs.
\begin{equation}
    \mathbf{y} = \softmax \left( \mathbf{W}_c \; \mathbf{e}_{\text{EE}} + \mathbf{b}_c \right),
\end{equation}
where $\mathbf{W}_c \in \mathbb{R}^{r \times d_z}$ and $\mathbf{b}_c \in \mathbb{R}^{r}$ are learned parameters of the classification layer and $r$ is the number of relation categories.

\section{Experimental Settings}

The model was developed using PyTorch~\cite{paszke2017automatic}. We incorporated early stopping to identify the best training epoch and used Adam~\cite{kingma2014adam} as the model optimiser.

%%% DATA
\subsection{Data and Task Settings}

We evaluated the proposed model on two datasets: \\
\textbf{CDR} (BioCreative V): The Chemical-Disease Reactions dataset was created by \citet{li2016biocreative} for document-level RE. It consists of $1,500$ PubMed abstracts, which are split into three equally sized sets for training, development and testing. The dataset was manually annotated with binary interactions between Chemical and Disease concepts.
For this dataset, we utilised PubMed pre-trained word embeddings~\cite{chiu2016train}.

\noindent \textbf{GDA} (DisGeNet): The Gene-Disease Associations dataset was introduced by \citet{wu2019renet}, containing $30,192$ MEDLINE abstracts, split into $29,192$ articles for training and $1,000$ for testing. The dataset was annotated with binary interactions between Gene and Disease concepts at the document-level, using distant supervision. 
Associations between concepts were generated by aligning the DisGeNet~\cite{pinero2016disgenet} platform with PubMed\footnote{\url{https://www.ncbi.nlm.nih.gov/pubmed/}} abstracts. 
We further split the training set into a 80/20 percentage split as training and development sets.
For the GDA dataset, we used randomly initialized word embeddings.

\subsection{Model Settings}
We explore multiple settings of the proposed graph using different edges (MM, ME, MS, ES, SS) and enhancements (node type embeddings, mention-pairs context embeddings, distance embeddings). We name our model EoG, an abbreviation of Edge-oriented Graph. 
We briefly describe the model settings in this section.

\noindent EoG refers to our main model with edges \{MM, ME, MS, ES, SS \}.
The EoG ({\tt Full}) setting refers to a model with a \textit{fully connected} graph, where the graph nodes are all connected to each other, including E nodes. For this purpose, we introduce an additional linear layer for the EE edges as in Equation~(\ref{eq:linear}). 
The EoG ({\tt NoInf}) setting refers to a \textit{no inference} model, where the iterative inference algorithm~(Section~\ref{sec:infenrence}) is ignored. The concatenation of the entity node embeddings is used to represent the target pair. In this case, we also make use of an additional EE linear layer for EE edges.
\begin{table*}[t!]
    \centering
    \scalebox{0.85}{
    \begin{tabular}{@{\extracolsep{9pt}}lccccccccc@{}}
        \toprule
        \multirow{2}{*}{Method} 
        & \multicolumn{3}{c}{Overall (\%)}
        & \multicolumn{3}{c}{Intra (\%)} 
        & \multicolumn{3}{c}{Inter (\%)} \\ 
        \cmidrule{2-4} \cmidrule{5-7} \cmidrule{8-10}
        & P & R & F1
        & P & R & F1 
        & P & R & F1 \\ \midrule
        
        \citet{gu2017chemical}     & 55.7 & 68.1 & 61.3
                                   & 59.7 & 55.0 & 57.2    
                                   & 51.9 & 7.0 & 11.7 \\
                                          
        \citet{verga2018simultaneously}   & 55.6 & 70.8 & 62.1 
                                          &-&-&- 
                                          &-&-&- \\
                                          
        \citet{nguyen2018convolutional}   & 57.0 & 68.6 & 62.3 
                                          &-&-&- 
                                          &-&-&- \\
                                          
        \midrule                        
        EoG     &  62.1 & 65.2   & \textbf{63.6}
                &  64.0	& 73.0   & \textbf{68.2} 
                &  56.0	& 46.7	 & \textbf{50.9} \\
                
        EoG ({\tt Full}) & 59.1  & 	56.2 & 	57.6
                         & 71.2  & 	62.3 & 	66.5
                         & 37.1  &  42.0 &  39.4 \\
                
        EoG ({\tt NoInf}) &  48.2 &	50.2   &  49.2
                          &  65.8 &	55.2   &  60.2
                          &  25.4 &	38.5   &  30.6 \\
                                 
        EoG ({\tt Sent})  & 56.9  &	53.5  &	55.2
                          &	56.9  &	76.4  &	65.2
                          & -     &  -    &- \\
                                
        \midrule \midrule
        
        \citet{zhou2016exploiting} & 55.6 & 68.4 & 61.3  
                                   &-&-&-
                                   &- &- &- \\
        
        \citet{peng2016improving}    & 62.1 & 64.2 & 63.1 
                                     &-&-&- 
                                     &-&-&- \\
                                     
        \citet{li2016cidextractor}   & 60.8 & 76.4 & 67.7
                                     & 67.3 & 52.4 & 58.9  
                                     &-&-&- \\
        
        \citet{panyam2018exploiting} & 53.2 & 69.7 & 60.3
                                     & 54.7 & 80.6 & 65.1  
                                     & 47.8 & 43.8 & 45.7 \\
                                
        \citet{zheng2018effective}    & 56.2 & 67.9 & 61.5
                                      &-&-&- 
                                      &-&-&- \\
        \bottomrule
    \end{tabular}
    }
    \caption{Overall, intra- and inter-sentence pairs performance comparison with the state-of-the-art on the CDR test set. The methods below the double line take advantage of additional training data and/or incorporate external tools.} 
    \label{tab:cdr_results}
\end{table*}
Finally, the EoG ({\tt Sent}) setting refers to a model that was trained on \textit{sentences} instead of documents. For each entity-level pair we merge the predictions of the mention-level pairs in different sentences using a maximum assumption: if at least one mention-level prediction indicates a relation then we predict the entity pair as related, similarly to \citet{gu2017chemical}.
All of the settings incorporate node type embeddings, contextual embeddings for MM edges and distance embeddings for MM and SS edges, unless otherwise stated.

%%% RESULTS
\section{Results}

% ===================  RESULTS =============== %
Table~\ref{tab:cdr_results} depicts the performance of our proposed model on the CDR test set, in comparison with the state-of-the-art.
We directly compare our model with models that do not incorporate external knowledge. \citet{verga2018simultaneously} and \citet{nguyen2018convolutional} consider a single pair per document, while \citet{gu2017chemical} develops separate models for intra- and inter-sentence pairs.
As it can be observed, the proposed model outperforms the state-of-the-art in CDR dataset by $1.3$ percentage points of overall performance. 
We also show the methods that take advantage of syntactic dependency tools. \citet{li2016cidextractor} uses co-training with additional unlabeled training data. 
Our model performs significantly better on intra- and inter-sentential pairs, even compared to most of the models with external knowledge, except for \citet{li2016cidextractor}.

In addition, we report the performance of three baseline models.
The EoG model outperforms all baselines for all pair types.
In particular, for the inter-sentence pairs, performance significantly drops with a fully connected graph ({\tt Full}) or without inference ({\tt NoInf}). The former might indicate the existence of certain reasoning paths that should be followed in order to relate entities residing in different sentences.
It is also important to note that the intra-sentence pairs substantially benefit from the document-level information, as EoG surpasses the performance of training on single sentences ({\tt Sent}) by $3$\%.
Finally, the performance drop in intra-sentence pairs, as a result of the inference algorithm removal ({\tt NoInf}), suggests that multiple entity associations exist in sentences~\cite{christopoulou2018walk}. Their interactions can be beneficial in cases of lack of word context information.

% ========= GDA ========= %
\begin{table}[t!]
    \centering
    \setlength{\tabcolsep}{4pt}
    \scalebox{0.85}{
    \begin{tabular}{l@{\hspace{3\tabcolsep}}c|c@{\hspace{4\tabcolsep}}c|c@{\hspace{4\tabcolsep}}c|c}
        \toprule
        \multirow{2}{*}{Model} & \multicolumn{6}{c}{Dev $\vert$ Test \ \ \ \ F1 (\%)} \\
        \cmidrule{2-7}
        & \multicolumn{2}{@{\hspace{-4\tabcolsep}}c}{Overall} 
        & \multicolumn{2}{@{\hspace{-4\tabcolsep}}c}{Intra}
        & \multicolumn{2}{@{\hspace{-1\tabcolsep}}c}{Inter} \\ \midrule
        %\citet{wu2019renet}        & 83.4 & 87.0 & 55.6  \\

        EoG               & 78.7 & 81.5   & 82.5 & 85.2   & 48.8 & 50.0  \\
        EoG ({\tt Full})  & 78.6 & 80.8   & 82.4 & 84.1   & 52.3 & 54.7  \\
        EoG ({\tt NoInf}) & 71.8 & 74.6   & 76.8 & 79.1   & 45.5 & 49.3  \\
        EoG ({\tt Sent})  & 73.8 & 73.8   & 78.1 & 78.8   & -    & -     \\
        \bottomrule
    \end{tabular}
    }
    \caption{Performance comparison on the GDA development and test sets.}
    \label{tab:gda_results}
\end{table}
We also apply our model on the distantly supervised GDA dataset. As shown in Table~\ref{tab:gda_results} results for intra-sentence pairs are consistent with the findings of the CDR dataset for both development and test sets. This indicates that  document-level information is helpful. However, performance differs for inter-sentence pairs and in particular for the fully connected graph ({\tt Full}) baseline. We partially attribute this behavior to the small number of inter-sentence pairs in the GDA dataset (only $13$\% compared to $30$\% in the CDR dataset) that results in inadequate learning patters for EoG. We leave further investigation as part of future work.

\section{Analysis \& Discussion}
We first analyse the performance of our main model (EoG) using different pre-trained word embeddings. Table~\ref{tab:embeddings} shows the performance difference between domain-specific ({\tt PubMed})~\cite{chiu2016train}, general-domain ({\tt GloVe})~\cite{pennington2014glove} and randomly initialized ({\tt random}) word embeddings. As observed, our proposed model performs consistently with both in-domain and out-of-domain pre-trained word embeddings. The low performance of random embeddings is due to the small size of the dataset, which results in lower quality embeddings.

\begin{table}[t!]
    \centering
    \setlength{\tabcolsep}{10pt}
    \scalebox{0.85}{
    \begin{tabular}{lrrr}
        \toprule
        \multirow{2}{*}{Embeddings} & \multicolumn{3}{c}{F1 (\%)} \\
        \cmidrule{2-4}
        & \multicolumn{1}{c}{Overall} 
        & \multicolumn{1}{c}{Intra}
        & \multicolumn{1}{c}{Inter} \\ \midrule 
        
        EoG ({\tt PubMed}) & 63.62 & 68.25 & 50.94  \\
        EoG ({\tt GloVe})  & 63.01 & 67.52 & 50.26 \\
        EoG ({\tt random}) & 61.41 & 66.80 & 46.51 \\
        \bottomrule
    \end{tabular}
    }
    \caption{Performance of EoG on the CDR test set with different pre-trained word embeddings.}
    \label{tab:embeddings}
\end{table}

\begin{figure}[t!]
    \centering
    \captionsetup[subfigure]{aboveskip=-1pt,belowskip=-1.5pt}
    \begin{subfigure}[b]{0.5\textwidth}
        \centering
        \includegraphics[width=0.83\textwidth]{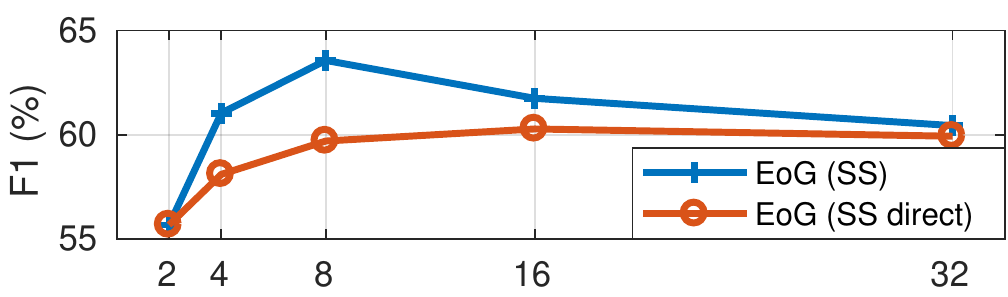}
        \caption{Overall}
        \label{fig:a}
    \end{subfigure}
    \begin{subfigure}[b]{0.5\textwidth}
        \centering
        \includegraphics[width=0.83\textwidth]{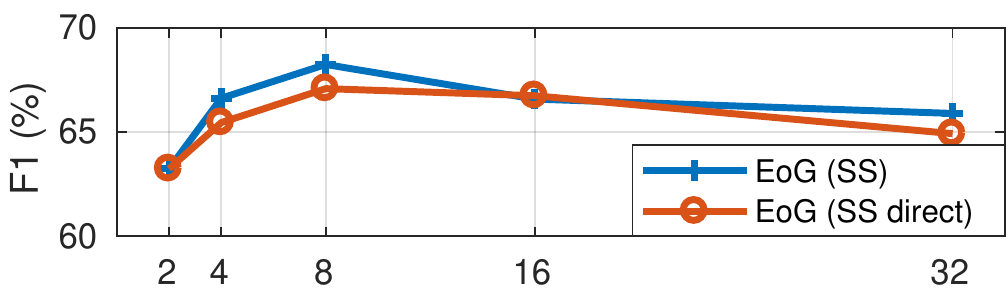}
        \caption{Intra-sentential}
        \label{fig:b}
    \end{subfigure}
    \begin{subfigure}[b]{0.5\textwidth}
        \centering
        \includegraphics[width=0.83\textwidth]{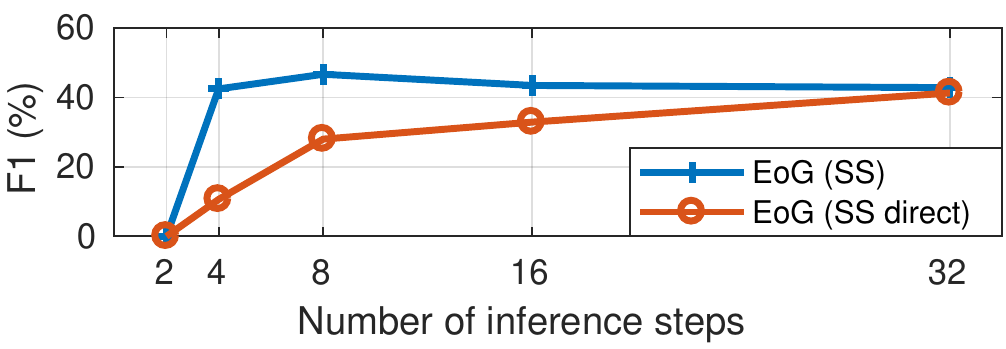}
        \caption{Inter-sentential}
        \label{fig:c}
    \end{subfigure}
    \caption{Performance as a function of the number of inference steps when using direct ($\text{SS}_\text{direct}$) or direct and indirect (SS) sentence-to-sentence edges, on the CDR development set.}
    \label{fig:seq_ind_edges}
\end{figure}

% ==================== ABLATION ================= %
For further analysis, we choose the CDR dataset as it is manually annotated.
To better analyse the behaviour of our model, we conduct analysis on the effect of direct and indirect sentence-to-sentence edges as a function of the inference steps. 
Figures~\ref{fig:a}, \ref{fig:b} and \ref{fig:c} illustrate the performance of both graphs for overall, intra- and inter-sentence pairs respectively.

The first observation is that usage of direct edges only, reduces the overall performance almost by $4\%$, for inference step $l=8$. 
This drop mostly affects inter-sentence pairs, where a $18\%$ point drop is observed. 
In fact, ordered edges ($\text{SS}_\text{direct}$) need longer inference to perform better, in comparison with additional indirect edges (SS) for which less steps are required. 
The superiority of SS edges, for all inference steps, compared to $\text{SS}_{\text{direct}}$ edges on inter-sentence pairs detection, indicates that in a narrative, some intermediate information is not important.
The observation that indirect edges perform slightly better than direct for intra-sentence pairs ($l \leq 16$) agrees with the results of Table~\ref{tab:cdr_results} where we showed that inter-sentence information can act as complementary evidence for intra-sentence pairs.

% ==================== CDR ablation edges ============= %
\begin{table}[t!]
    \centering
    \setlength{\tabcolsep}{10pt}
    \scalebox{0.85}{
    \begin{tabular}{lrrr}
        \toprule
        \multirow{2}{*}{Edge Types} & \multicolumn{3}{c}{F1 (\%)} \\
        \cmidrule{2-4}
        & \multicolumn{1}{c}{Overall} 
        & \multicolumn{1}{c}{Intra}
        & \multicolumn{1}{c}{Inter} \\ \midrule 
        
        {\tt EE}        & 55.14 & 61.31 & 40.34 \\
        \midrule
        
        EoG                                      & 63.57  & 68.25  & 46.68 \\
        \ \ {\tt $-$MM}                          & 62.77  & 67.93  & 46.65 \\ 
        \ \ {\tt $-$ME}                          & 61.57  & 66.39  & 45.40 \\
        \ \ {\tt $-$MS}                          & 62.92  & 67.55  & 44.74 \\ 
        \ \ {\tt $-$ES}                          & 61.41  & 66.44  & 43.04 \\
        \ \ {\tt $-\text{SS}_{\text{indirect}}$} & 59.70  & 67.09  & 28.00 \\
        \ \ {\tt $-$SS}                          & 57.41  & 65.45  & 1.59 \\
        \midrule
        
        \ \ {\tt $-$MM,ME,MS}  & 60.46  & 66.07  & 39.56 \\ 
        \ \ {\tt $-$ES,MS,SS}  & 56.86  & 64.63  & 0.00  \\
        \bottomrule
    \end{tabular}
    }
    \caption{Ablation analysis for different edge and node types  on the CDR development set.}
    \label{tab:cdr_abl_edge}
\end{table}

We additionally conduct ablation analysis on the graph edges and nodes, as shown in Table~\ref{tab:cdr_abl_edge}. 
Usage of EE edges only results in poor performance across pairs. 
Removal of MM and ME edges does not significantly affect the performance as ES edges can replace their impact. Complete removal of connections to M nodes results in low inter-sentence performance.
This behaviour pinpoints the importance of some local dependencies in identifying cross-sentence relations. 

\begin{figure}[t!]
    %\centering
    \begin{subfigure}[t]{0.24\textwidth}
        \centering
        \includegraphics[scale=1.5]{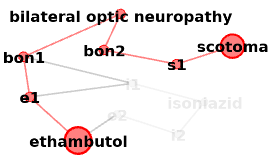}
        \caption{MM, ME edges}
        \label{fig:ex_paths_a}
    \end{subfigure}%
    \begin{subfigure}[t]{0.24\textwidth}
        \centering
        \includegraphics[scale=1.5]{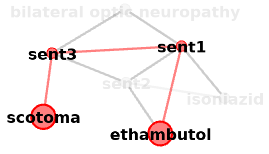}
        \caption{ES, SS edges}
        \label{fig:ex_paths_b}
    \end{subfigure}
    \caption{Relation paths with different types of edges.}
    \label{fig:ex_paths}
\end{figure}

Removal of ES edges reduces the performance of all pairs, as encoding of EE edges becomes more difficult\footnote{Length $3$ (E-M-M-E) for intra- and length $5$ (E-M-S-S-M-E) for inter-sentence pairs.}.
We further observe very poor identification of inter-sentence pairs without sentence-to-sentence connections. 
This is complementary with the inability of the model to identify any inter-sentence pairs without connections to S nodes. 
In this scenario, we enable identification of pairs across sentences only through MM and ME edges, as shown in Figure~\ref{fig:ex_paths_a}. In the CDR dataset, $78$\% of inter-sentential pairs have at least one argument that is mentioned only once in the document. The identification of these pairs, without S nodes, requires very long inference paths\footnote{Minimum inference length $6$ (E-M-M-E-M-M-E).}. As shown in Figure~\ref{fig:ex_paths_b}, the introduction of S nodes results in a path with half the length, which we expect to better represent the relation.
Longer inference representations are much weaker than shorter ones.
This suggests that the inference mechanism has limited capability in identifying very complex associations.

% ==================== Ablation settings ================ %
% CDR ablation settings
\begin{table}[t!]
    \centering
    \scalebox{0.85}{
    \begin{tabular}{lccc}
        \toprule
        \multirow{2}{*}{Model} & \multicolumn{3}{c}{F1 (\%)} \\
        \cmidrule{2-4}
        & Overall & Intra & Inter  \\
        \midrule
        EoG                             & 63.57   & 68.25 & 46.68 \\
        \ \ {\tt $-$node types (T)}     & 62.31   & 67.50 & 44.80 \\
        \ \ {\tt $-$MM context (C)}     & 62.88   & 67.67 & 46.59 \\
        \ \ {\tt $-$distances (D)}      & 62.53   & 68.00 & 41.53 \\
        \ \ {\tt $-$T,C,D}              & 63.10   & 68.44 & 43.48 \\
        \bottomrule
    \end{tabular}
    }
    \caption{Ablation analysis of edge enhancements on the CDR development set.}
    \label{tab:cdr_abl_set}
\end{table}

We then investigate the additional enhancements of the graph edges in Table~\ref{tab:cdr_abl_set}.
In general, intra-sentence pairs are not affected by these settings. However, for inter-sentence pairs, removal of node type embeddings and distance embeddings results in a $2\%$ and $5\%$ drop in terms of F1-score. These results indicate that the interactions between different elements in a document, along with the distance between sentences and mentions, play an important role in inter-sentence pair inference.  
Removing all of these settings does not perform worse than removing one of them, which might indicate model overfitting. We plan to further investigate this as part of future work.

\begin{figure}[t!]
    \centering
    \includegraphics[width=0.45\textwidth]{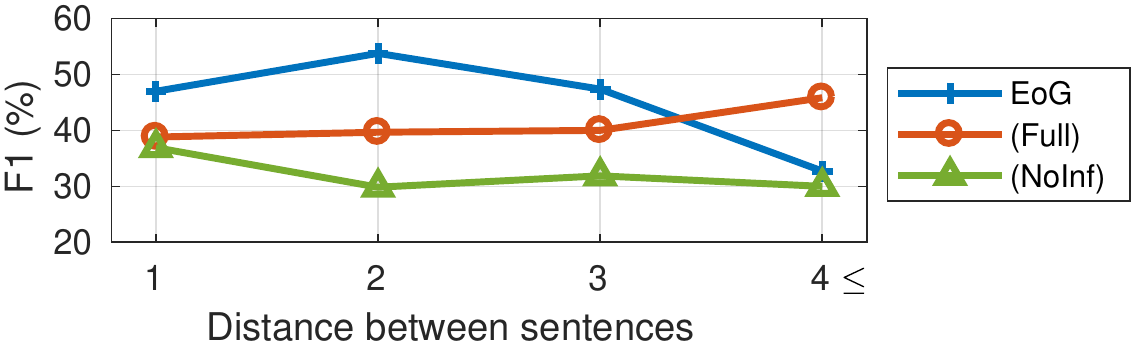}
    \caption{Performance of inter-sentence pairs on the CDR development set as a function of their sentence distance.}
    \label{fig:dist}
\end{figure}

We examine the performance of different models on inter-sentence pairs, based on their sentence-level distances. Figure~\ref{fig:dist} illustrates that for long-distanced pairs, EoG has lower performance, indicating the difficulty in predicting them and a possible requirement for other, latent document-level information (EoG ({\tt Full})).

\begin{table}[t!]
    \centering
    \scalebox{0.85}{
    \begin{tabularx}{0.5\textwidth}{X}
        \hline
        Following short exposure to oral \textbf{prednisone} [...]. Both presented
        in the emergency room with profound \textbf{coma}, \textbf{hypotension}, severe \textbf{hyperglycemia},
        and acidosis. \\
        \hline
        The etiology of \textbf{pyeloureteritis cystica} has long been [...] The \textit{disease} occurred subsequent to the initiation of \textbf{heparin} therapy [...] \\
        \hline
        Time trends in \textbf{warfarin}-associated {\textit{\textbf{hemorrhage}}}. [...] The proportion
        of patients with major and {\textit{\textbf{intracranial bleeding}}} increased [...] \\
        \hline
    \end{tabularx}
    }
    \caption{Inter-sentence pairs from the CDR development set that EoG fails to detect.}
    \label{tab:errors}
\end{table}

As final analysis, we investigate some of the cases where the graph models are unable to identify inter-sentence related pairs. For this purpose, we randomly check some of the common false negative errors among the EoG models. We identify three frequent cases of errors, as shown in Table~\ref{tab:errors}.
In the first case, when multiple entities reside in the same sentence and are connected with conjunctions (e.g., `and') or commas, the model often failed to find associations with all of them. The second error derives from missing co-reference connections. For instance, \textit{pyeloureteritis cystica} is referred to as \textit{disease}. Although our model cannot directly create these edges, S nodes potentially simulate such links, by encoding the co-referring entities into the sentence representation. Finally, incomplete entity linking results into additional model errors. For instance, in the third example, \textit{hemorrhage} and \textit{intracranial bleeding} are synonymous terms. However, they are assigned different KB IDs, hence treated as different entities. The model can find the intra-sentential relation but not the inter-sentential one.

%%% RELATED WORK
\section{Related Work}
Traditional approaches focus on intra-sentence supervised RE, utilising CNN or RNN, ignoring multiple entities in a sentence~\cite{zeng2014relation,nguyen2015relation} as well as incorporating external syntactic tools~\cite{miwa2016end,zhang2018graph}. 
\citet{christopoulou2018walk} considered intra-sentence entity interactions without domain dependencies by modelling long dependencies between the entities of a sentence. 

Other approaches deal with distantly-supervised datasets but are also limited to intra-sentential relations. They utilise Piecewise Convolutional Neural Networks (PCNN)~\cite{zeng2015distant}, attention mechanisms~\cite{lin2016neural,zhou2018distant}, entity descriptors~\cite{jiang2016relation} and graph CNNs~\cite{vashishth2018reside} to perform MIL on bags-of-sentences that contain multiple mentions of an entity pair. Recently, \citet{zeng2017paths} proposed a method for extracting paths between entities using the target entities' mentions in several different sentences (in possibly different documents) as intermediate connectors. They allow mention-mention edges only if these mentions belong to the same entity and consider that a single mention pair exists in a sentence. On the contrary, we not only allow interactions between all mentions in the same sentence, but also consider multiple edges between mentions, entities and sentences in a document. 

Current approaches that try to deal with document-level RE are mostly graph-based. 
\citet{quirk2017distant} introduced the notion of a document graph, where nodes are words and edges represent intra- and inter-sentential relations between the words. They connected words with different dependency edges and trained a binary logistic regression classifier. They evaluated their model on distantly supervised full-text articles from PubMed for Gene-Drug associations, restricting pairs within a window of consecutive sentences.
Following this work, other approaches incorporated graphical models for document-level RE such as graph LSTM~\cite{peng2017cross}, graph CNN~\cite{song2018graph} or RNNs on dependency tree structures~\cite{gupta2019neural}.
Recently, \citet{jia2019document} improved $n$-ary RE using information from multiple sentences and paragraphs in a document. Similar to our approach, they choose to directly classify concept-level pairs rather than multiple mention-level pairs. Although they consider sub-relations to model related tuples, they ignore interactions with other entities outside of the target tuple in the discourse units. 

Non-graph-based approaches utilise different intra- and inter-sentence models and merge the resulted predictions~\cite{gu2016chemical,gu2017chemical}. Other approaches extract document-level representations for each candidate entity pair~\cite{zheng2018effective,li2018chemical,wu2019renet}, or use syntactic dependency structures~\cite{zhou2016exploiting,peng2016improving}.
\citet{verga2018simultaneously} proposed a Transformer-based model for document-level relation extraction with multi-instance learning, merging multiple mention pairs. 
\citet{nguyen2018convolutional} used a CNN with additional character-level embeddings.
\citet{singh2019relation} also utilised Transformer and connected two target entities by combining them directly and via a contextual token. However, they consider a single target entity pair per document.

%%% CONCLUSION
\section{Conclusion}
We presented a novel edge-oriented graph neural model for document-level relation extraction using multi-instance learning. 
The proposed model constructs a document-level graph with heterogeneous types of nodes and edges, modelling intra- and inter-sentence pairs simultaneously with an iterative algorithm over the graph edges. To the best of our knowledge, this is the first approach to utilise an edge-oriented model for document-level RE.

Analysis on intra- and inter-sentence pairs indicated that the proposed, partially-connected, document graph structure can effectively encode dependencies between document elements. Additionally, we deduce that document-level information can contribute to the identification of intra-sentence pairs leading to higher precision and F1-score.

As future work, we plan to improve the inference mechanism and potentially incorporate additional information in the document-graph structure.
We hope that this study will inspire the community to further investigate the usage of edge-oriented models on RE and other related tasks.

\section*{Acknowledgments}
The authors would like to thank the anonymous reviewers for their comments and AIST/AIRC for providing the computational resources for this study. Research was funded by the University of Manchester James Elson Studentship Award and the BBSRC Japan Partnering Award BB/P025684/1.

\bibliographystyle{acl_natbib}
\bibliography{biblio}

\begin{thebibliography}{44}
\expandafter\ifx\csname natexlab\endcsname\relax\def\natexlab#1{#1}\fi

\bibitem[{Banko et~al.(2007)Banko, Cafarella, Soderland, Broadhead, and
  Etzioni}]{banko2007open}
Michele Banko, Michael~J Cafarella, Stephen Soderland, Matt Broadhead, and Oren
  Etzioni. 2007.
\newblock Open information extraction from the web.
\newblock In \emph{Proceedings of the International Joint Conference on
  Artifical Intelligence}, pages 2670--2676. Morgan Kaufmann Publishers Inc.

\bibitem[{Borgwardt and Kriegel(2005)}]{Borgwardt2005shortest}
Karsten~M Borgwardt and Hans-Peter Kriegel. 2005.
\newblock Shortest-path kernels on graphs.
\newblock In \emph{Proceedings of the IEEE International Conference on Data
  Mining}, pages 74--81. IEEE Computer Society.

\bibitem[{Chiu et~al.(2016)Chiu, Crichton, Korhonen, and
  Pyysalo}]{chiu2016train}
Billy Chiu, Gamal Crichton, Anna Korhonen, and Sampo Pyysalo. 2016.
\newblock How to train good word embeddings for biomedical nlp.
\newblock In \emph{Proceedings of the {B}io{NLP} workshop}, pages 166--174.

\bibitem[{Christopoulou et~al.(2018)Christopoulou, Miwa, and
  Ananiadou}]{christopoulou2018walk}
Fenia Christopoulou, Makoto Miwa, and Sophia Ananiadou. 2018.
\newblock A walk-based model on entity graphs for relation extraction.
\newblock In \emph{Proceedings of the Annual Meeting of the Association for
  Computational Linguistics (Volume 2)}, pages 81--88. Association for
  Computational Linguistics.

\bibitem[{Gu et~al.(2016)Gu, Qian, and Zhou}]{gu2016chemical}
Jinghang Gu, Longhua Qian, and Guodong Zhou. 2016.
\newblock Chemical-induced disease relation extraction with various linguistic
  features.
\newblock \emph{Database}.

\bibitem[{Gu et~al.(2017)Gu, Sun, Qian, and Zhou}]{gu2017chemical}
Jinghang Gu, Fuqing Sun, Longhua Qian, and Guodong Zhou. 2017.
\newblock Chemical-induced disease relation extraction via convolutional neural
  network.
\newblock \emph{Database}.

\bibitem[{Gupta et~al.(2019)Gupta, Rajaram, Schütze, and
  Runkler}]{gupta2019neural}
Pankaj Gupta, Subburam Rajaram, Hinrich Schütze, and Thomas Runkler. 2019.
\newblock Neural relation extraction within and across sentence boundaries.
\newblock In \emph{Proceedings of the AAAI Conference on Artificial
  Intelligence}, volume~33, pages 6513--6520.

\bibitem[{Hochreiter and Schmidhuber(1997)}]{hochreiter1997long}
Sepp Hochreiter and J{\"u}rgen Schmidhuber. 1997.
\newblock Long short-term memory.
\newblock \emph{Neural computation}, 9(8):1735--1780.

\bibitem[{Jia et~al.(2019)Jia, Wong, and Poon}]{jia2019document}
Robin Jia, Cliff Wong, and Hoifung Poon. 2019.
\newblock Document-level n-ary relation extraction with multiscale
  representation learning.
\newblock In \emph{Proceedings of the Conference of the North {A}merican
  Chapter of the Association for Computational Linguistics: Human Language
  Technologies (Volume 1)}, pages 3693--3704. Association for Computational
  Linguistics.

\bibitem[{Jiang et~al.(2016)Jiang, Wang, Li, and Wang}]{jiang2016relation}
Xiaotian Jiang, Quan Wang, Peng Li, and Bin Wang. 2016.
\newblock Relation extraction with multi-instance multi-label convolutional
  neural networks.
\newblock In \emph{Proceedings of the International Conference on Computational
  Linguistics: Technical Papers}, pages 1471--1480. The COLING Organizing
  Committee.

\bibitem[{Kingma and Ba(2015)}]{kingma2014adam}
Diederik~P Kingma and Jimmy Ba. 2015.
\newblock Adam: A method for stochastic optimization.
\newblock In \emph{Proceedings of the International Conference on Learning
  Representations}.

\bibitem[{Li et~al.(2018)Li, Yang, Chen, Tang, Wang, and Yan}]{li2018chemical}
Haodi Li, Ming Yang, Qingcai Chen, Buzhou Tang, Xiaolong Wang, and Jun Yan.
  2018.
\newblock Chemical-induced disease extraction via recurrent piecewise
  convolutional neural networks.
\newblock \emph{BMC medical informatics and decision making}, 18(2):60.

\bibitem[{Li et~al.(2016{\natexlab{a}})Li, Sun, Johnson, Sciaky, Wei, Leaman,
  Davis, Mattingly, Wiegers, and Lu}]{li2016biocreative}
Jiao Li, Yueping Sun, Robin~J Johnson, Daniela Sciaky, Chih-Hsuan Wei, Robert
  Leaman, Allan~Peter Davis, Carolyn~J Mattingly, Thomas~C Wiegers, and Zhiyong
  Lu. 2016{\natexlab{a}}.
\newblock Biocreative v cdr task corpus: a resource for chemical disease
  relation extraction.
\newblock \emph{Database}.

\bibitem[{Li et~al.(2016{\natexlab{b}})Li, Yang, Lin, Wang, Gui, Zhang, and
  Wang}]{li2016cidextractor}
Zhiheng Li, Zhihao Yang, Hongfei Lin, Jian Wang, Yingyi Gui, Yin Zhang, and Lei
  Wang. 2016{\natexlab{b}}.
\newblock {CID}extractor: A chemical-induced disease relation extraction system
  for biomedical literature.
\newblock In \emph{IEEE International Conference on Bioinformatics and
  Biomedicine}, pages 994--1001. IEEE.

\bibitem[{Lin et~al.(2016)Lin, Shen, Liu, Luan, and Sun}]{lin2016neural}
Yankai Lin, Shiqi Shen, Zhiyuan Liu, Huanbo Luan, and Maosong Sun. 2016.
\newblock Neural relation extraction with selective attention over instances.
\newblock In \emph{Proceedings of the Annual Meeting of the Association for
  Computational Linguistics (Volume 1)}, pages 2124--2133.

\bibitem[{Luan et~al.(2019)Luan, Wadden, He, Shah, Ostendorf, and
  Hajishirzi}]{luan2019general}
Yi~Luan, Dave Wadden, Luheng He, Amy Shah, Mari Ostendorf, and Hannaneh
  Hajishirzi. 2019.
\newblock A general framework for information extraction using dynamic span
  graphs.
\newblock In \emph{Proceedings of the Conference of the North {A}merican
  Chapter of the Association for Computational Linguistics: Human Language
  Technologies (Volume 1)}, pages 3036--3046. Association for Computational
  Linguistics.

\bibitem[{Mintz et~al.(2009)Mintz, Bills, Snow, and
  Jurafsky}]{mintz2009distant}
Mike Mintz, Steven Bills, Rion Snow, and Dan Jurafsky. 2009.
\newblock Distant supervision for relation extraction without labeled data.
\newblock In \emph{Proceedings of the Joint Conference of the Annual Meeting of
  the ACL and the 4th International Joint Conference on Natural Language
  Processing of the AFNLP (Volume 2)}, pages 1003--1011. Association for
  Computational Linguistics.

\bibitem[{Miwa and Bansal(2016)}]{miwa2016end}
Makoto Miwa and Mohit Bansal. 2016.
\newblock End-to-end relation extraction using {LSTM}s on sequences and tree
  structures.
\newblock In \emph{Proceedings of the Annual Meeting of the Association for
  Computational Linguistics (Volume 1)}, pages 1105--1116. Association for
  Computational Linguistics.

\bibitem[{Nguyen and Verspoor(2018)}]{nguyen2018convolutional}
Dat~Quoc Nguyen and Karin Verspoor. 2018.
\newblock Convolutional neural networks for chemical-disease relation
  extraction are improved with character-based word embeddings.
\newblock In \emph{Proceedings of the {B}io{NLP} workshop}, pages 129--136.
  Association for Computational Linguistics.

\bibitem[{Nguyen and Grishman(2015)}]{nguyen2015relation}
Thien~Huu Nguyen and Ralph Grishman. 2015.
\newblock Relation extraction: Perspective from convolutional neural networks.
\newblock In \emph{Proceedings of the Workshop on Vector Space Modeling for
  Natural Language Processing}, pages 39--48.

\bibitem[{Panyam et~al.(2018)Panyam, Verspoor, Cohn, and
  Ramamohanarao}]{panyam2018exploiting}
Nagesh~C Panyam, Karin Verspoor, Trevor Cohn, and Kotagiri Ramamohanarao. 2018.
\newblock Exploiting graph kernels for high performance biomedical relation
  extraction.
\newblock \emph{Journal of biomedical semantics}, 9(1):7.

\bibitem[{Paszke et~al.(2017)Paszke, Gross, Chintala, Chanan, Yang, DeVito,
  Lin, Desmaison, Antiga, and Lerer}]{paszke2017automatic}
Adam Paszke, Sam Gross, Soumith Chintala, Gregory Chanan, Edward Yang, Zachary
  DeVito, Zeming Lin, Alban Desmaison, Luca Antiga, and Adam Lerer. 2017.
\newblock Automatic differentiation in pytorch.
\newblock In \emph{NIPS-W}.

\bibitem[{Peng et~al.(2017)Peng, Poon, Quirk, Toutanova, and
  Yih}]{peng2017cross}
Nanyun Peng, Hoifung Poon, Chris Quirk, Kristina Toutanova, and Wen-tau Yih.
  2017.
\newblock Cross-sentence n-ary relation extraction with graph lstms.
\newblock \emph{Transactions of the Association for Computational Linguistics},
  5:101--115.

\bibitem[{Peng et~al.(2016)Peng, Wei, and Lu}]{peng2016improving}
Yifan Peng, Chih-Hsuan Wei, and Zhiyong Lu. 2016.
\newblock Improving chemical disease relation extraction with rich features and
  weakly labeled data.
\newblock \emph{Journal of cheminformatics}, 8(1):53.

\bibitem[{Pennington et~al.(2014)Pennington, Socher, and
  Manning}]{pennington2014glove}
Jeffrey Pennington, Richard Socher, and Christopher~D Manning. 2014.
\newblock Glove: Global vectors for word representation.
\newblock In \emph{Proceedings of the Empirical Methods in Natural Language
  Processing}, pages 1532--1543. Association for Computational Linguistics.

\bibitem[{Pi{\~n}ero et~al.(2016)Pi{\~n}ero, Bravo, Queralt-Rosinach,
  Guti{\'e}rrez-Sacrist{\'a}n, Deu-Pons, Centeno, Garc{\'\i}a-Garc{\'\i}a,
  Sanz, and Furlong}]{pinero2016disgenet}
Janet Pi{\~n}ero, {\`A}lex Bravo, N{\'u}ria Queralt-Rosinach, Alba
  Guti{\'e}rrez-Sacrist{\'a}n, Jordi Deu-Pons, Emilio Centeno, Javier
  Garc{\'\i}a-Garc{\'\i}a, Ferran Sanz, and Laura~I Furlong. 2016.
\newblock Disgenet: a comprehensive platform integrating information on human
  disease-associated genes and variants.
\newblock \emph{Nucleic acids research}, pages D833--D839.

\bibitem[{Quirk and Poon(2017)}]{quirk2017distant}
Chris Quirk and Hoifung Poon. 2017.
\newblock Distant supervision for relation extraction beyond the sentence
  boundary.
\newblock In \emph{Proceedings of the Conference of the European Chapter of the
  Association for Computational Linguistics (Volume 1)}, pages 1171--1182.
  Association for Computational Linguistics.

\bibitem[{Riedel et~al.(2010)Riedel, Yao, and McCallum}]{riedel2010modeling}
Sebastian Riedel, Limin Yao, and Andrew McCallum. 2010.
\newblock Modeling relations and their mentions without labeled text.
\newblock In \emph{Joint European Conference on Machine Learning and Knowledge
  Discovery in Databases}, pages 148--163. Springer.

\bibitem[{Schuster and Paliwal(1997)}]{schuster1997bidirectional}
Mike Schuster and Kuldip~K Paliwal. 1997.
\newblock Bidirectional recurrent neural networks.
\newblock \emph{IEEE Transactions on Signal Processing}, 45(11):2673--2681.

\bibitem[{Shen et~al.(2014)Shen, Wang, and Han}]{shen2014entity}
Wei Shen, Jianyong Wang, and Jiawei Han. 2014.
\newblock Entity linking with a knowledge base: Issues, techniques, and
  solutions.
\newblock \emph{IEEE Transactions on Knowledge and Data Engineering},
  27(2):443--460.

\bibitem[{Singh and Bhatia(2019)}]{singh2019relation}
Gaurav Singh and Parminder Bhatia. 2019.
\newblock Relation extraction using explicit context conditioning.
\newblock In \emph{Proceedings of the Conference of the North {A}merican
  Chapter of the Association for Computational Linguistics: Human Language
  Technologies, Volume 1 (Long and Short Papers)}, pages 1442--1447.
  Association for Computational Linguistics.

\bibitem[{Song et~al.(2018)Song, Zhang, Wang, and Gildea}]{song2018graph}
Linfeng Song, Yue Zhang, Zhiguo Wang, and Daniel Gildea. 2018.
\newblock N-ary relation extraction using graph-state lstm.
\newblock In \emph{Proceedings of the Conference on Empirical Methods in
  Natural Language Processing}, pages 2226--2235. Association for Computational
  Linguistics.

\bibitem[{Vashishth et~al.(2018)Vashishth, Joshi, Prayaga, Bhattacharyya, and
  Talukdar}]{vashishth2018reside}
Shikhar Vashishth, Rishabh Joshi, Sai~Suman Prayaga, Chiranjib Bhattacharyya,
  and Partha Talukdar. 2018.
\newblock Reside: Improving distantly-supervised neural relation extraction
  using side information.
\newblock In \emph{Proceedings of the Conference on Empirical Methods in
  Natural Language Processing}, pages 1257--1266. Association for Computational
  Linguistics.

\bibitem[{Verga et~al.(2018)Verga, Strubell, and
  McCallum}]{verga2018simultaneously}
Patrick Verga, Emma Strubell, and Andrew McCallum. 2018.
\newblock Simultaneously self-attending to all mentions for full-abstract
  biological relation extraction.
\newblock In \emph{Proceedings of the Conference of the North American Chapter
  of the Association for Computational Linguistics: Human Language Technologies
  (Volume 1)}, pages 872--884. Association for Computational Linguistics.

\bibitem[{Wang et~al.(2016)Wang, Cao, de~Melo, and Liu}]{wang2016relation}
Linlin Wang, Zhu Cao, Gerard de~Melo, and Zhiyuan Liu. 2016.
\newblock Relation classification via multi-level attention {CNN}s.
\newblock In \emph{Proceedings of the Annual Meeting of the Association for
  Computational Linguistics (Volume 1)}, pages 1298--1307. Association for
  Computational Linguistics.

\bibitem[{Wu et~al.(2019)Wu, Luo, Leung, Ting, and Lam}]{wu2019renet}
Ye~Wu, Ruibang Luo, Henry~CM Leung, Hing-Fung Ting, and Tak-Wah Lam. 2019.
\newblock Renet: A deep learning approach for extracting gene-disease
  associations from literature.
\newblock In \emph{International Conference on Research in Computational
  Molecular Biology}, pages 272--284. Springer.

\bibitem[{Xu et~al.(2015)Xu, Mou, Li, Chen, Peng, and Jin}]{xu2015classifying}
Yan Xu, Lili Mou, Ge~Li, Yunchuan Chen, Hao Peng, and Zhi Jin. 2015.
\newblock Classifying relations via long short term memory networks along
  shortest dependency paths.
\newblock In \emph{Proceedings of Conference on Empirical Methods in Natural
  Language Processing}, pages 1785--1794. Association for Computational
  Linguistics.

\bibitem[{Zeng et~al.(2015)Zeng, Liu, Chen, and Zhao}]{zeng2015distant}
Daojian Zeng, Kang Liu, Yubo Chen, and Jun Zhao. 2015.
\newblock Distant supervision for relation extraction via piecewise
  convolutional neural networks.
\newblock In \emph{Proceedings of the Conference on Empirical Methods in
  Natural Language Processing}, pages 1753--1762.

\bibitem[{Zeng et~al.(2014)Zeng, Liu, Lai, Zhou, and Zhao}]{zeng2014relation}
Daojian Zeng, Kang Liu, Siwei Lai, Guangyou Zhou, and Jun Zhao. 2014.
\newblock Relation classification via convolutional deep neural network.
\newblock In \emph{Proceedings of the International Conference on Computational
  Linguistics: Technical Papers}, pages 2335--2344. Dublin City University and
  Association for Computational Linguistics.

\bibitem[{Zeng et~al.(2017)Zeng, Lin, Liu, and Sun}]{zeng2017paths}
Wenyuan Zeng, Yankai Lin, Zhiyuan Liu, and Maosong Sun. 2017.
\newblock Incorporating relation paths in neural relation extraction.
\newblock In \emph{Proceedings of the Conference on Empirical Methods in
  Natural Language Processing}, pages 1768--1777. Association for Computational
  Linguistics.

\bibitem[{Zhang et~al.(2018)Zhang, Qi, and Manning}]{zhang2018graph}
Yuhao Zhang, Peng Qi, and Christopher~D Manning. 2018.
\newblock Graph convolution over pruned dependency trees improves relation
  extraction.
\newblock In \emph{Proceedings of the Conference on Empirical Methods in
  Natural Language Processing}, pages 2205--2215. Association for Computational
  Linguistics.

\bibitem[{Zheng et~al.(2018)Zheng, Lin, Li, Liu, Li, Xu, Zhang, Yang, and
  Wang}]{zheng2018effective}
Wei Zheng, Hongfei Lin, Zhiheng Li, Xiaoxia Liu, Zhengguang Li, Bo~Xu, Yijia
  Zhang, Zhihao Yang, and Jian Wang. 2018.
\newblock An effective neural model extracting document level chemical-induced
  disease relations from biomedical literature.
\newblock \emph{Journal of biomedical informatics}, 83:1--9.

\bibitem[{Zhou et~al.(2016)Zhou, Deng, Chen, Yang, Jia, and
  Huang}]{zhou2016exploiting}
Huiwei Zhou, Huijie Deng, Long Chen, Yunlong Yang, Chen Jia, and Degen Huang.
  2016.
\newblock Exploiting syntactic and semantics information for chemical--disease
  relation extraction.
\newblock \emph{Database}.

\bibitem[{Zhou et~al.(2018)Zhou, Xu, Qi, Bao, Chen, and Xu}]{zhou2018distant}
Peng Zhou, Jiaming Xu, Zhenyu Qi, Hongyun Bao, Zhineng Chen, and Bo~Xu. 2018.
\newblock Distant supervision for relation extraction with hierarchical
  selective attention.
\newblock \emph{Neural Networks}, 108:240--247.

\end{thebibliography}

\appendix

\section{Datasets}

In Tables~\ref{tab:cdr_statistcs}-\ref{tab:gda_statistcs} we summarise the statistics for the CDR and GDA datasets, respectively.
For all datasets, we used the GENIA Sentence Splitter\footnote{\url{http://www.nactem.ac.uk/y-matsu/geniass/}} and GENIA Tagger\footnote{\url{http://www.nactem.ac.uk/GENIA/tagger/}} for sentence splitting and word tokenisation respectively.
We additionally removed mentions in the given abstracts that were not grounded to a Knowledge Base ID (ID equal to $-1$).

Due to the small size of the CDR dataset, some approaches create a new split from the union of train and development sets~\cite{verga2018simultaneously,zhou2018distant}. We select to merge the train and development sets and re-train our model on their entire union for evaluation on the test set following \citet{lin2016neural} and \citet{zhou2016exploiting}.
To compare with related work, we followed \citet{verga2018simultaneously} and \citet{gu2016chemical} and ignored non-related pairs that correspond to general concepts (MeSH vocabulary hypernym filtering).

\begin{table}[ht!]
    \centering
    \scalebox{0.85}{
    \begin{tabular}{lrrr}
        \hline
                        & Train & Dev & Test \\
        \hline
        Documents       &  500  & 500  & 500 \\
        
        Positive pairs  & 1,038  & 1,012 & 1,066 \\
        \multicolumn{1}{r}{Intra}   & 754   & 766  & 747 \\
        \multicolumn{1}{r}{Inter}   & 284   & 246  & 319 \\
        
        Negative pairs  & 4,202  & 4,075 & 4,138 \\
        
        Entities        & \\ 
        \multicolumn{1}{r}{Chemical}  & 1,467 & 1,507 & 1,434 \\
        \multicolumn{1}{r}{Disease}   & 1,965 & 1,864 &	1,988 \\
       
        Mentions        & \\ 
        \multicolumn{1}{r}{Chemical}  & 5,162 &	5,307 &	5,370 \\
        \multicolumn{1}{r}{Disease}   & 4,252 &	4,328 &	4,430 \\
        \hline
    \end{tabular}
    }
    \caption{CDR (BioCreative V) dataset statistics.}
    \label{tab:cdr_statistcs}
\end{table}

\begin{table}[ht!]
    \centering
    \scalebox{0.85}{
    \begin{tabular}{lccc}
        \hline
        & Train & Dev & Test \\
        \hline
        Documents       &  23,353 &	5,839 &	1,000 \\
        
        Positive pairs                & 36,079 &	8,762 &	1,502 \\
        \multicolumn{1}{r}{Intra}     & 30,905 &	7,558 &	1,305 \\
        \multicolumn{1}{r}{Inter}     & 5,174  &	1,204 &	197 \\
        
        Negative pairs  & 96,399 &	24,362 &	3,720 \\ 
        
        Entities        & \\
        \multicolumn{1}{r}{Gene}        & 46,151 &	11,406 &	1,903 \\
        \multicolumn{1}{r}{Disease}     & 67,257 &	16,703 &	2,778 \\ 
       
        Mentions        & \\
        \multicolumn{1}{r}{Gene}        & 205,457 &	51,410 &	8,404 \\
        \multicolumn{1}{r}{Disease}     & 226,015 &	56,318 &	9,524 \\
        \hline
    \end{tabular}
    }
    \caption{GDA (DisGeNet) dataset statistics.}
    \label{tab:gda_statistcs}
\end{table}

%\newpage
\section{Hyper-parameter Setting}
We used the development set to identify the stopping training epoch and tune the number of inference iterations. 
Except from these parameters, all experiments used the same hyperparameters, with a fixed initialisation seed.
For the CDR dataset EoG, ({\tt Full}) and ({\tt Sent}) models performed best with $l=8, 2, 4$ inference steps, respectively. The chosen batchsize was equal to $2$. 
For the GDA dataset, EoG and EoG ({\tt Full}) performed best with $l=16$ and EoG ({\tt Sent}) with $l=8$ inference steps. 
The chosen batchsize was equal to $3$.
For all experiments performance was measured in terms of micro precision (P), recall (R) and F1-score (F1).
We list the hyper-parameters used to train the proposed model in Table~\ref{tab:hyperparams}.

\begin{table}[ht!]
    \centering
    \scalebox{0.85}{
    \begin{tabular}{lr}
    \hline
        Parameter & Value  \\
        \hline
        Batch size                      & $[2, 3]$ \\
        Learning rate                   & $0.002$ \\
        Gradient clipping               & $10$ \\
        Early stop patience             & $10$ \\
        Regularization                  & $10^{-4}$ \\
        Dropout word embedding layer    & $0.5$ \\
        Dropout classification layer    & $0.3$ \\
        Word dimension                  & $200$ \\
        Node type dimension             & $10$ \\
        Distance dimension              & $10$ \\
        Edge dimension                  & $100$ \\
        $\beta$                         & $0.8$ \\
        Optimizer                       & Adam \\
        Inference iterations            & $[0, 5]$ \\
        \hline
    \end{tabular}
    }
    \caption{Hyper-parameter values used in the reported experiments.}
    \label{tab:hyperparams}
\end{table}

\end{document}